\documentclass{llncs}
\usepackage{times}
\usepackage{latexsym}
\usepackage{multirow}
\usepackage{amsmath,amsfonts}
\usepackage{array}
\usepackage{breqn}
\usepackage{enumerate}
\usepackage{tikz}
\usepackage{diagbox}
\usepackage{algorithm}% http://ctan.org/pkg/algorithms
\usepackage{algpseudocode}% http://ctan.org/pkg/algorithmicx
\usepackage{float}
\usepackage{url}
\usepackage{caption}

\newcolumntype{C}[1]{>{\centering\let\newline\\\arraybackslash\hspace{0pt}}m{#1}}

\begin{document}

\pagestyle{headings}

\title{Neural Multi-Step Reasoning for Question Answering on Semi-Structured Tables}
	
	\author{Till Haug, Octavian-Eugen Ganea and Paulina Grnarova}
	
	\authorrunning{Haug et al.}
	
	\institute{Department of Computer Science \\ ETH Zurich, Switzerland\\till@veezoo.com, octavian.ganea@inf.ethz.ch, paulina.grnarova@inf.ethz.ch}

\maketitle

%%%%%%%%%%%%%%%%%%%%%%%%%%%%%%%%%%%%%%%%%%%%%%%%%%%%%%%%%%%%%%%%%%%%%%%%%%%%%%%%%%%%%
%\vspace{-0.5cm}
\begin{abstract}
We explore neural network models for answering multi-step reasoning questions that operate on semi-structured tables. Challenges arise from deep logical compositionality and domain openness. Our approach is weakly supervised, trained on question-answer-table triples. It generates human readable \textit{logical forms} from natural language questions, which are then ranked based on word and character convolutional neural networks. A model ensemble achieved at the moment of publication state-of-the-art score on the WikiTableQuestions dataset.
\end{abstract}

%%%%%%%%%%%%%%%%%%%%%%%%%%%%%%%%%%%%%%%%%%%%%%%%%%%%%%%%%%%%%%%%%%%%%%%%%%%%%%%%%%%%%
%\vspace{-0.5cm}
\section{Introduction}
%\vspace{-0.2cm}
Teaching computers to answer complex natural language questions  requires sophisticated reasoning and human language understanding. We investigate generic natural language interfaces for simple arithmetic questions on semi-structured tables. Typical questions for this task are topic independent and may require performing multiple discrete operations such as aggregation, comparison, superlatives or arithmetics. 

We propose a weakly supervised neural model that eliminates the need for expensive feature engineering in the candidate ranking stage. Each natural language question is translated using the method of~\cite{pasupat2015compositional} into a set of machine understandable candidate representations, called \textit{logical forms} or \textit{programs}. Then, the most likely such program is retrieved in two steps: i) using a simple algorithm, logical forms are transformed back into \textit{paraphrases} (textual representations) understandable by non-expert users, ii) next, these strings are further embedded together with their respective questions in a jointly learned vector space using convolutional neural networks over character and word embeddings. Multi-layer neural networks and bilinear mappings are further employed as effective similarity measures and combined to score the candidate interpretations. Finally, the highest ranked logical form is executed against the input data to retrieve the answer. Our method uses only weak-supervision from question-answer-table input triples, without requiring expensive annotations of gold logical forms. 

We empirically test our approach on a series of experiments on WikiTableQuestions, to our knowledge the only dataset designed for this task. An ensemble of our best models reached state-of-the-art accuracy of 38.7\% at the moment of publication. 
%\footnote{\url{http://nlp.stanford.edu/software/sempre/wikitable/viewer/}}

%%%%%%%%%%%%%%%%%%%%%%%%%%%%%%%%%%%%%%%%%%%%%%%%%%%%%%%%%%%%%%%%%%%%%%%%%%%%%%%%%%%%%
%\vspace{-0.2cm}
\section{Related Work}
%\vspace{-0.2cm}
%In the context of QA with semi-structured knowledge,~\cite{khashabi2016question} use integer linear programming.
We briefly mention here two main types of QA systems related to our task\footnote{An extensive list of open-domain QA publications can be found here: \url{https://aclweb.org/aclwiki/Question_Answering_(State_of_the_art)}}: semantic parsing-based and embedding-based. \textit{Semantic parsing-based} methods perform a functional parse of the question that is further converted to a machine understandable program and executed on a knowledgebase or database. For QA on semi-structured tables with multi-compositional queries, \cite{pasupat2015compositional} generate and rank candidate logical forms with a log-linear model, resorting to hand-crafted features for scoring. As opposed, we learn neural features for each question and the paraphrase of each candidate logical form. Paraphrases and hand-crafted features have successfully facilitated semantic parsers targeting simple factoid \cite{berant2014semantic} and compositional questions \cite{wang2015building}. Compositional questions are also the focus of \cite{neelakantan2016learning} that construct logical forms from the question embedding through operations parametrized by RNNs, thus losing interpretability. A similar fully neural, end-to-end differentiable network was proposed by \cite{yin2015neural}.

\textit{Embedding-based} methods determine compatibility between a question-answer pair using embeddings in a shared vector space~\cite{bordes2014question}. Embedding learning using deep learning architectures has been widely explored in other domains, e.g. in the context of sentiment classification~\cite{kim2014convolutional}.

%\vspace{-0.3cm}

\section{Model}
%\vspace{-0.3cm}

We describe our QA system. For every question $q$ : i) a set of candidate logical forms $\{z_i\}_{i = 1, \ldots , n_q}$ is generated using the method of~\cite{pasupat2015compositional}; ii) each such candidate program $z_i$ is transformed in an interpretable textual representation $t_i$ ; iii) all $t_i$'s are jointly embedded with $q$ in the same vector space and scored using a neural similarity function; iv)  the logical form $z_i^*$ corresponding to the highest ranked $t_i^*$ is selected as the machine-understandable translation of question $q$ and executed on the input table to retrieve the final answer. Our contributions are the novel models that perform steps ii) and iii), while for step i) we rely on the work of~\cite{pasupat2015compositional} (henceforth: PL2015).

%\vspace{-0.4cm}
\subsection{Candidate Logical Form Generation}\label{clfg}
%\vspace{-0.2cm}

We generate a set of candidate logical forms from a question using the method of~\cite{pasupat2015compositional}. Only briefly, we review this method. Specifically, a question is parsed into a set of candidate logical forms using a semantic parser that recursively applies deduction rules. Logical forms are represented in Lambda DCS form~\cite{liang2013lambda} and can be executed on a table to yield an answer. An example of a question and its correct logical form are below:\\
\texttt{How many people attended the last Rolling Stones concert?}\\
\texttt{\textbf{R}[$\lambda x$[Attendance.Number.$x$]].argmax(Act.RollingStones,Index)}.

%\vspace{-0.4cm}
\subsection{Converting Logical Forms to Text}
%\vspace{-0.2cm}
In Algorithm 1 we describe how logical forms are transformed into interpretable textual representations called "paraphrases". We choose to embed paraphrases in low dimensional vectors and compare these against the question embedding. Working directly with paraphrases instead of logical forms is a design choice, justified by their interpretability, comprehensibility (understandability by non-technical users) and empirical accuracy gains. Our method recursively traverses the tree representation of the logical form starting at the root. For example, the correct candidate logical form for the question mentioned in section~\ref{clfg}, namely \texttt{How many people attended the last Rolling Stones concert?}, is mapped to the paraphrase \texttt{Attendance as number of last table row where act is Rolling Stones}.

%\vspace{-0.4cm}

\subsection{Joint Embedding Model} \label{sec:models}
%\vspace{-0.2cm}
We embed the question together with the paraphrases of candidate logical forms in a jointly learned vector space.  We use two  convolutional neural networks (CNNs) for question and paraphrase embeddings, on top of which a max-pooling operation is applied. The CNNs receive as input token embeddings obtained as described below.

%\vspace{-0.4cm}

\algnewcommand\algorithmicswitch{\textbf{switch}}
\algnewcommand\algorithmiccase{\textbf{case}}
\algnewcommand\algorithmicassert{\texttt{assert}}
\algnewcommand\Assert[1]{\State \algorithmicassert(#1)}%
% New "environments"
\algdef{SE}[SWITCH]{Switch}{EndSwitch}[1]{\algorithmicswitch\ #1\ \algorithmicdo}{\algorithmicend\ \algorithmicswitch}%
\algdef{SE}[CASE]{Case}{EndCase}[1]{\algorithmiccase\ #1}{\algorithmicend\ \algorithmiccase}%
\algtext*{EndSwitch}%
\algtext*{EndCase}%

\begin{algorithm*}
%\tiny
\caption{Recursive paraphrasing of a Lambda DCS logical form. The + operation means string concatenation with spaces. Lambda DCS language is detailed in \protect\cite{liang2013lambda}.}
\begin{algorithmic}[1]
\Procedure{Paraphrase}{$z$} \Comment{$z$ is the root of a Lambda DCS logical form}
\Switch{$z$}
    \Case{Aggregation} \Comment{e.g. count, max, min...}
      \State $t\gets \textsc{Aggregation}(z) + \textsc{Paraphrase}(z.child)$
    \EndCase
    \Case{Join} \Comment{join on relations, e.g. $\lambda x$.Country($x$, Australia)}
      \State $t\gets\textsc{Paraphrase}(z.relation)$ + $\textsc{Paraphrase}(z.child)$
    \EndCase
        \Case{Reverse} \Comment{reverses a binary relation}
      \State $t\gets\textsc{Paraphrase}(z.child)$
    \EndCase
        \Case{LambdaFormula} \Comment{lambda expression $\lambda x.[...]$}
      \State $t\gets\textsc{Paraphrase}(z.body)$
    \EndCase
        \Case{Arithmetic \textbf{or} Merge} \Comment{e.g. plus, minus, union...}
      \State $t\gets\textsc{Paraphrase}(z.left) + \textsc{Operation}(z) + \textsc{Paraphrase}(z.right)$
    \EndCase
        \Case{Superlative} \Comment{e.g. argmax(x, value)}
      \State $t\gets\textsc{Operation}(z) + \textsc{Paraphrase}(z.value) + \textsc{Paraphrase}(z.relation)$
    \EndCase
        \Case{Value} \Comment{i.e. constants}
      \State $t\gets z.value$
    \EndCase
  \EndSwitch
    \State \textbf{return} $t$ \Comment{$t$ is the textual paraphrase of the Lambda DCS logical form}
\EndProcedure
\end{algorithmic}
\end{algorithm*}

%\vspace{-1.3cm}
\subsubsection{Token Embedding}
%\begin{figure*}[t]
%\centering
%\includegraphics[width=1\textwidth]{embeddings}
%\caption{Conversion of a sentence into a token embedding matrix. Fig. inspired from \protect\cite{kim2015character}.}
%\label{fig:tokenembedding}
%\end{figure*}

\begin{figure}[!tbp]
\hspace{-0.7cm}
%\centering
    \includegraphics[width=1.1\textwidth]{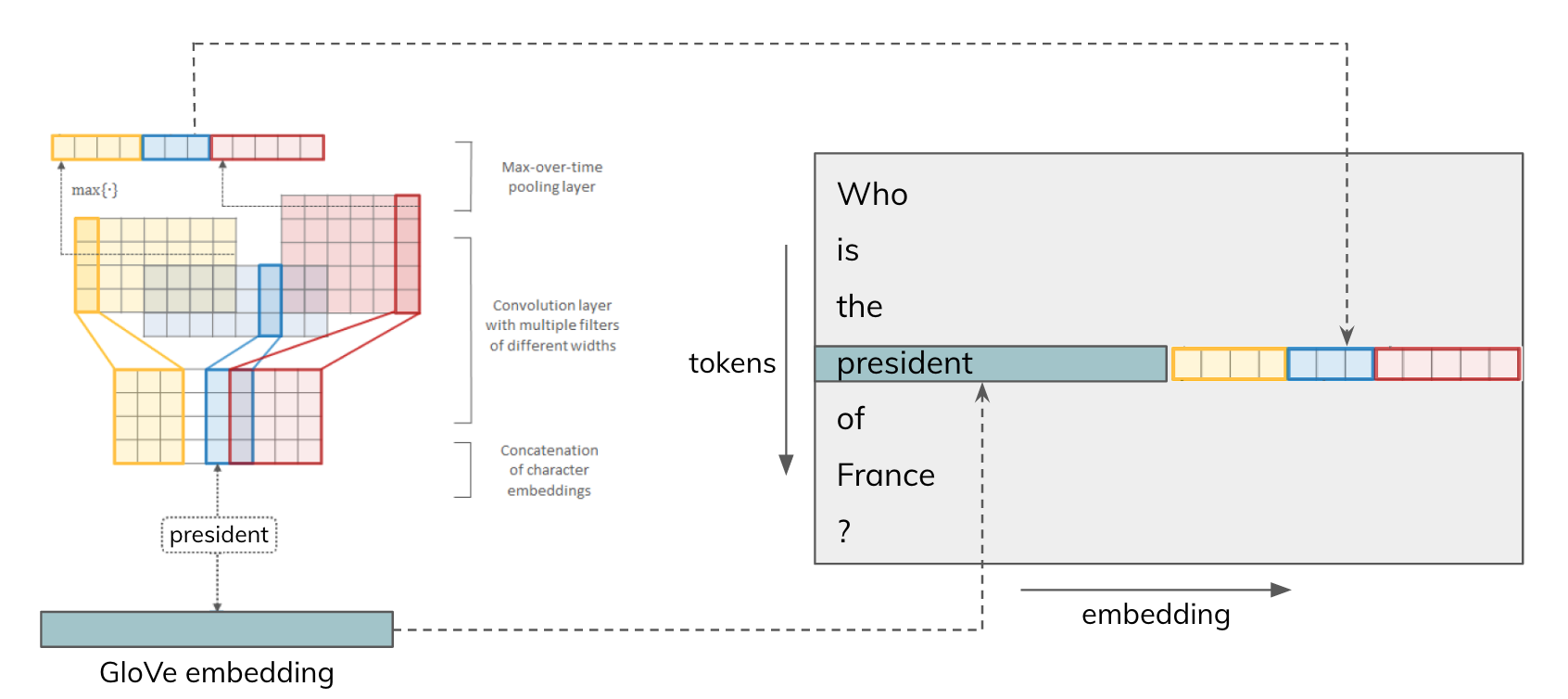}
    \caption{Conversion of a sentence into a token embedding matrix using word embeddings and char CNNs. The resulting matrix is again fed to a CNN/RNN  (not pictured) to produce a sentence embedding. Fig. inspired from \protect\cite{kim2015character}.}
    \label{fig:tokenembedding}
%\vspace{-0.5cm}
\end{figure}

The embedding of an input word sequence (e.g. question, paraphrase) is depicted in Figure~\ref{fig:tokenembedding} and is similar to~\cite{kim2015character}. Every token is parametrized by learnable word and character embeddings. The latter help dealing with unknown tokens (e.g. rare words, misspellings, numbers or dates). Token vectors are then obtained using a CNN (with multiple filter widths) over the constituent characters , followed by a max-over-time pooling layer and concatenation with the word vector. 

%\vspace{-0.4cm}
\subsubsection{Sentence Embedding} 
We map both the question $q$ and the paraphrase $t$ into a joint vector space using sentence embeddings obtained from two jointly trained CNNs. CNNs' filters span a different number of tokens from a width set $L$. For each filter width $l \in L$, we learn $n$ different filters, each of dimension $\mathbb{R}^{l\times d}$, where $d$ is the word embedding size. After the convolution layer, we apply a max-over-time pooling on the resulting feature matrices which yields, per filter-width, a vector of dimension $n$. 
Next, we concatenate the resulting max-over-time pooling vectors of the different filter-widths in $L$ to form our sentence embedding. The final sentence embedding size is $n|L|$.

%\vspace{-0.3cm}
\subsubsection{Neural Similarity Measures} 
%\begin{figure}[t]
%\centering
%\includegraphics[width=0.3\textwidth]{single_arch}
%\caption{Our best single model, CNN-FC-BILIN. }
%\label{fig:variant}
%\end{figure}

Let $u,v \in \mathbb{R}^{d}$ be the sentence embeddings of question $q$ and of paraphrase $t$. We experiment with the following similarity scores:
i) \textsc{DOTPRODUCT} : $u^{T}v$ ; ii) \textsc{BILIN} : $u^{T}Sv$ , with $S\in\mathbb{R}^{d\times d}$ being a trainable matrix; iii) \textsc{FC}: u and v concatenated, followed by two sequential fully connected layers with ELU non-linearities; iv) \textsc{FC-BILIN}: weighted average of BILIN and FC. These models define  parametrized similarity scoring functions $\varPhi: Q\times T\rightarrow\mathbb{R}$, where $Q$ is the set of natural language questions and $T$ is the set of paraphrases of logical forms. 

%\begin{figure}[!tbp]
%%  \centering
%  \hspace{-0.2cm}
%  \begin{minipage}[b]{0.7\textwidth}
%    \includegraphics[width=\textwidth]{embeddings}
%    \caption{Conversion of a sentence into a token embedding matrix. Fig. inspired from \protect\cite{kim2015character}.}
%    \label{fig:tokenembedding}
%  \end{minipage}
%  \hfill
%  \hspace{0.1cm}
%  \begin{minipage}[b]{0.29\textwidth}
%    \includegraphics[width=\textwidth]{single_arch}
%    \caption{Our best single model, CNN-FC-BILIN.}
%    \label{fig:variant}
%  \end{minipage}
%\vspace{-1cm}
%\end{figure}

%\vspace{-0.3cm}
\subsection{Training Algorithm}

%\vspace{-0.2cm}

For training, we build two sets $\mathcal{P}$ (positive) and $\mathcal{N}$ (negative) consisting of all pairs $(q,t) \in Q \times T$ of questions and paraphrases of candidate logical forms generated as described in Section~\ref{clfg}. A pair is positive or negative if its logical form gives the correct or respectively incorrect gold answer when executed on the corresponding table. During training, we use the ranking hinge loss function (with margin $\theta$):
\begin{align*}
{L(\mathcal{P},\mathcal{N})= \sum_{p\in\mathcal{P}}\sum_{n\in\mathcal{N}}\max(0,}\theta-\varPhi(p)+\varPhi(n))
\end{align*}
%\vspace{-1cm}

\section{Experiments}
%\vspace{-0.2cm}

\textbf{Dataset: }
For training and testing we use the train-validation-test split of WikiTableQuestions~\cite{pasupat2015compositional}, a dataset containing 22,033 pairs of questions and answers based on 2,108 Wikipedia tables. This dataset is also used by our  baselines,~\cite{pasupat2015compositional,neelakantan2016learning}. Tables are not shared across these splits, which requires models to generalize to unseen data. We obtain about 3.8 million training  triples $(q,t,l)$, where $l$ is a binary indicator of whether the logical form gives the correct gold answer when executed on the corresponding table. 76.7\% of the questions have at least one correct candidate logical form when generated with the model of~\cite{pasupat2015compositional}.

\begin{table}
\setlength{\tabcolsep}{5pt}
\small
\begin{tabular}{cc}
\begin{tabular}{|c|c|}
\hline
\textbf{Baseline Systems} & \textbf{P@1}\tabularnewline
\hline 
\begin{tabular}{l@{}c@{}}Neural Programmer~\cite{neelakantan2016learning} \\ (single model) \end{tabular} & 34.2\%\tabularnewline
\hline 
\begin{tabular}{l@{}c@{}}Neural Programmer~\cite{neelakantan2016learning} \\(15 ensemble models)\end{tabular} & 37.7\%\tabularnewline
\hline 
PL2015~\cite{pasupat2015compositional} & 37.1\%\tabularnewline
\hline 
\end{tabular}& 
\begin{tabular}{|c|c|}
\hline
\textbf{Our Models} & \textbf{P@1}\tabularnewline
\hline 
CNN-DOTPRODUCT & 31.4\%\tabularnewline
\hline 
CNN-BILIN & 20.4\%\tabularnewline
\hline 
CNN-FC & 30.4\%\tabularnewline
\hline 
RNN-FC-BILIN & 29.6\%\tabularnewline
\hline 
CNN-FC-BILIN (best single model) & 34.8\%\tabularnewline
\hline 
CNN-FC-BILIN (15 ensemble models) & \textbf{38.7\%}\tabularnewline
\hline 
\end{tabular}
\end{tabular}
\caption{Precision@1 of various baselines and our models on the WikiTableQuestions dataset. }
\label{tab:Results}
%xxxxx\vspace{-1.2cm}
\end{table}

\begin{table*}
%\vspace{-0.5cm}
\begin{tabular}{c|c}
\textbf{Question} & \textbf{Paraphrase ({\color{green}gold} vs {\color{red}predicted})}\tabularnewline
\hline 
\multirow{2}{*}{{\footnotesize{}Which association entered last?}} & \textcolor{red}{\footnotesize{}association of last row}\tabularnewline
\cline{2-2} 
 & \textcolor{green}{\footnotesize{}association of row with highest number of joining
year}\tabularnewline
\hline 

\multirow{2}{*}{{\footnotesize{}What is the total of all the medals?}} & \textcolor{red}{\footnotesize{}count all rows}\tabularnewline
\cline{2-2} 
 & \textcolor{green}{\footnotesize{}number of total of nation is total}\tabularnewline
\hline 

\multirow{2}{*}{{\footnotesize{}\begin{tabular}{@{}c@{}}How many episodes were originally \\ aired before December
1965?\end{tabular}}} & \textcolor{red}{\footnotesize{}count original air date as date \textless= 12 1965}\tabularnewline
\cline{2-2} 
 & \textcolor{green}{\footnotesize{}count original air date as date \textless 12 1965}\tabularnewline
\end{tabular}
%\vspace{0.5cm}
\caption{Example of common errors of our model. \label{tab:Qualit}}
%\vspace{-1cm}
\end{table*}

\textbf{Training Details: }
Our models are implemented using TensorFlow and trained on a single Tesla P100 GPU. Training takes approximately 6 hours. We initialize word vectors with 200 dimensional GloVe (\cite{pennington2014glove}) pre-trained vectors. For the character CNN we use widths spanning 1, 2 and 3 characters. The sentence embedding CNNs use widths of $L=\{2,4,6,8\}$. The fully connected layers in the FC models have 500 hidden neurons, which we regularize using 0.8-dropout. The loss margin $\theta$ is set to 0.2. Optimization is done using Adam~\cite{kingma2014adam} with  a learning rate of 7e-4. Hyperparameters are tunned on the development data split of the Wiki-TableQuestions table. We choose the best performing model on the validation set using early stopping.

\textbf{Results: }
Experimental results are shown in Table~\ref{tab:Results}. Our best performing single model is FC-BILIN with CNNs,  Intuitively, BILIN and FC are able to extract different interaction features between the two input vectors, while their linear combination retains the best of both models. An ensemble of 15 single CNN-FC-BILIN models was setting (at the moment of publication) a new state-of-the-art precision@1 for this dataset: 38.7\%. This shows that the same model initialized differently can learn different features. We also experimented with recurrent neural networks (RNNs) for the sentence embedding since these are known to capture word order better than CNNs. However, RNN-FC-BILIN performs worse than its CNN variant. 

There are a few reasons that contributed to the low accuracy obtained on this task by various methods (including ours) compared to other NLP problems: weak supervision, small training size and a high percentage of unanswerable questions. 

%%%%%%%%%%%%%%%%%%%%%%%%%%%

%\vspace{-0.5cm}

\textbf{Error Analysis: }
The questions our models do not answer correctly can be split into two categories: either a correct logical form is not generated, or our scoring models do not rank the correct one at the top. We perform a qualitative analysis presented in Table~\ref{tab:Qualit} to reveal common question types our models often rank incorrectly. The first two examples show questions whose correct logical form depends on the structure of the table. In these cases a bias towards the more general logical form is often exhibited. The third example shows that our model has difficulty distinguishing operands with slight modification (e.g. smaller and smaller equals), which may be due to weak-supervision. 

\begin{table}
%\vspace{-0.5cm}
\setlength{\tabcolsep}{7pt}
\small
\begin{tabular}{cc}
\begin{tabular}{l|c}
\textbf{System} & \textbf{P@1} \tabularnewline
\hline 
CNN-FC-BILIN & 34.1\%\tabularnewline
\hline 
\qquad  w/o Dropout & 33.3\%\tabularnewline
\hline 
\qquad  w/o Char Embeddings & 33.8\%\tabularnewline
\hline 
\qquad  w/o GloVe (random init) & 32.4\%\tabularnewline
\hline 
\qquad  w/o Paraphrasing & 33.1\%\tabularnewline
\hline 
%\caption{Component contributions to our model. \label{tab:Ablation}}
\end{tabular} & 
\begin{tabular}{l|c}
\textbf{System} & \textbf{Amount}\tabularnewline
\hline 
Lookup & 10.8\%\tabularnewline
\hline 
\begin{tabular}{l@{}c@{}}Aggregation \& \\ Next/Previous \end{tabular} & 39.8\%\tabularnewline
\hline 
Superlatives & 30.1\%\tabularnewline
\hline 
Arithmetic \& Comparisons & 19.3\%\tabularnewline
\hline 
\end{tabular} %\caption{Types of correctly answered questions.   \label{tab:Distribution}}
\end{tabular}
%\vspace{0.1cm}
\caption{Ablation studies. Left: Component contributions to our model. Right: Types of questions answered correctly by our system. }
%\vspace{-1cm}
\label{tab:Ablation}
\end{table}

\textbf{Ablation Studies: }
For a better understanding of our model, we investigate the usefulness of various components with an ablation study shown in Table~\ref{tab:Ablation}. In particular, we emphasize that replacing the paraphrasing stage with the raw strings of the Lambda DCS expressions resulted in lower precision@1, which confirms the utility of this stage. 

\textbf{Analysis of Correct Answers: }
We analyze how well our best single model performs on various question types. For this, we manually annotate 80 randomly chosen questions that are correctly answered by our model and report statistics in Table \ref{tab:Ablation}.

%\vspace{-0.5cm}
\section{Conclusion}
%\vspace{-0.3cm}
In this paper we propose a neural network QA system for semi-structured tables that eliminates the need for manually designed features. Experiments show that an ensemble of our models reaches competitive accuracy on the WikiTableQuestions dataset, thus indicating its capability to answer complex, multi-compositional questions. Our code is available at \url{https://github.com/dalab/neural_qa} . % \\

\section*{Acknowledgments}
This research was supported by the Swiss National Science Foundation (SNSF) grant number 407540\_167176 under the project "Conversational Agent for Interactive Access to Information".

%%%%%%%%%%%%%%%%%%%%%%%%%%%%%%%%%%%%%%%%%%%%%%%%%%%%%%%%%%%%%%%%%%%%%%%%%%%%%%%%%%%%%
%\vspace{-0.3cm}
\bibliography{ecir}

\begin{thebibliography}{10}

\bibitem{berant2014semantic}
Jonathan Berant and Percy Liang.
\newblock Semantic parsing via paraphrasing.
\newblock In {\em ACL (1)}, pages 1415--1425, 2014.

\bibitem{bordes2014question}
Antoine Bordes, Sumit Chopra, and Jason Weston.
\newblock Question answering with subgraph embeddings.
\newblock {\em arXiv preprint arXiv:1406.3676}, 2014.

\bibitem{kim2014convolutional}
Yoon Kim.
\newblock Convolutional neural networks for sentence classification.
\newblock In {\em In EMNLP}. Citeseer, 2014.

\bibitem{kim2015character}
Yoon Kim, Yacine Jernite, David Sontag, and Alexander~M Rush.
\newblock Character-aware neural language models.
\newblock In {\em Proceedings of the Thirtieth AAAI Conference on Artificial
  Intelligence}, pages 2741--2749. AAAI Press, 2016.

\bibitem{kingma2014adam}
Diederik Kingma and Jimmy Ba.
\newblock Adam: A method for stochastic optimization.
\newblock {\em arXiv preprint arXiv:1412.6980}, 2014.

\bibitem{liang2013lambda}
Percy Liang.
\newblock Lambda dependency-based compositional semantics.
\newblock {\em arXiv:1309.4408}, 2013.

\bibitem{neelakantan2016learning}
Arvind Neelakantan, Quoc~V Le, Martin Abadi, Andrew McCallum, and Dario Amodei.
\newblock Learning a natural language interface with neural programmer.
\newblock {\em arXiv:1611.08945}, 2016.

\bibitem{pasupat2015compositional}
Panupong Pasupat and Percy Liang.
\newblock Compositional semantic parsing on semi-structured tables.
\newblock In {\em In Proceedings of the Annual Meeting of the Association for
  Computational Linguistics}. Citeseer, 2015.

\bibitem{pennington2014glove}
Jeffrey Pennington, Richard Socher, and Christopher~D Manning.
\newblock Glove: Global vectors for word representation.
\newblock In {\em EMNLP}, volume~14, pages 1532--1543, 2014.

\bibitem{wang2015building}
Yushi Wang, Jonathan Berant, Percy Liang, et~al.
\newblock Building a semantic parser overnight.
\newblock In {\em ACL (1)}, pages 1332--1342, 2015.

\bibitem{yin2015neural}
Pengcheng Yin, Zhengdong Lu, Hang Li, and Ben Kao.
\newblock Neural enquirer: learning to query tables in natural language.
\newblock In {\em Proceedings of the Twenty-Fifth International Joint
  Conference on Artificial Intelligence}, pages 2308--2314. AAAI Press, 2016.

\end{thebibliography}
\bibliographystyle{plain}

%%%%%%%%%%%%%%%%%%%%%%%%%%%%%%%%%%%%%%%%%%%%%%%%%%%%%%%%%%%%%%%%%%%%%%%%%%%%%%%%%%%%%

\end{document}